%% file: Main.tex
\documentclass[conference]{IEEEtran}
\IEEEoverridecommandlockouts

\usepackage{cite}
\usepackage{amsmath,amssymb,amsfonts}
\usepackage{algorithmic}
\usepackage{algorithm}
\usepackage{graphicx}
\usepackage{textcomp}
\usepackage{xcolor}
\usepackage{multirow}

\def\BibTeX{{\rm B\kern-.05em{\sc i\kern-.025em b}\kern-.08em
    T\kern-.1667em\lower.7ex\hbox{E}\kern-.125emX}}

\begin{document}

\title{Pruning Everything, Everywhere, All at Once}

\author{Gustavo Henrique do Nascimento, Ian Pons, Anna Helena Reali Costa and Artur Jordao
\\
Escola Politécnica de Engenharia, Universidade de São Paulo, Brazil
}

\maketitle

\input{Sections/Abstract}

\input{Sections/Introduction}
\input{Sections/RelatedWork}
\input{Sections/Methodology}
\input{Sections/Experiments}

\input{Sections/Conclusions}
\input{Sections/Acknowledgments}

\bibliographystyle{IEEEtran}
\bibliography{refs}

\input{Sections/Appendix}

\end{document}

%% file: Sections/Abstract.tex
\begin{abstract} 
Deep learning stands as the modern paradigm for solving cognitive tasks. However, as the problem complexity increases, models grow deeper and computationally prohibitive, hindering advancements in real-world and resource-constrained applications. Extensive studies reveal that pruning structures in these models efficiently reduces model complexity and improves computational efficiency. Successful strategies in this sphere include removing neurons (i.e., filters, heads) or layers, but not both together. Therefore, simultaneously pruning different structures remains an open problem. To fill this gap and leverage the benefits of eliminating neurons and layers at once, we propose a new method capable of pruning different structures within a model as follows. Given two candidate subnetworks (pruned models), one from layer pruning and the other from neuron pruning, our method decides which to choose by selecting the one with the highest representation similarity to its parent (the network that generates the subnetworks) using the Centered Kernel Alignment (CKA) metric. Iteratively repeating this process provides highly sparse models that preserve the original predictive ability. Throughout extensive experiments on standard architectures and benchmarks, we confirm the effectiveness of our approach and show that it outperforms state-of-the-art layer and filter pruning techniques. At high levels of Floating Point Operations (FLOPs) reduction, most state-of-the-art methods degrade accuracy, whereas our approach either improves it or experiences only a minimal drop. Notably, on the popular ResNet56 and ResNet110, we achieve a milestone of 86.37\% and 95.82\% FLOPs reduction. Besides, our pruned models obtain robustness to adversarial and out-of-distribution samples and take an important step towards GreenAI, reducing carbon emissions by up to 83.31\%. Overall, we believe our work opens a new chapter in pruning. Code is available at: https://github.com/NascimentoG/PruningEverything.
\end{abstract}

%% file: Sections/Introduction.tex
\section{Introduction}\label{sec:introduction}
Deep learning drives unprecedented progress in various cognitive tasks and serves as the powerhouse for learning patterns from data~\cite{Block:2025,Dubey:2024}. However, this performance incurs a significant computational cost, and consequently environmental impacts~\cite{Faiz:2024}. The recent LLama 3 model family comprehends a concrete example, where training requires up to 16K H100 GPUs globally distributed. Notably, the largest model (405B) requires 16 GPUs with 16-bit precision for inference~\cite{Dubey:2024}.

\begin{figure*}[!t]
\centering
\includegraphics[width=\columnwidth]{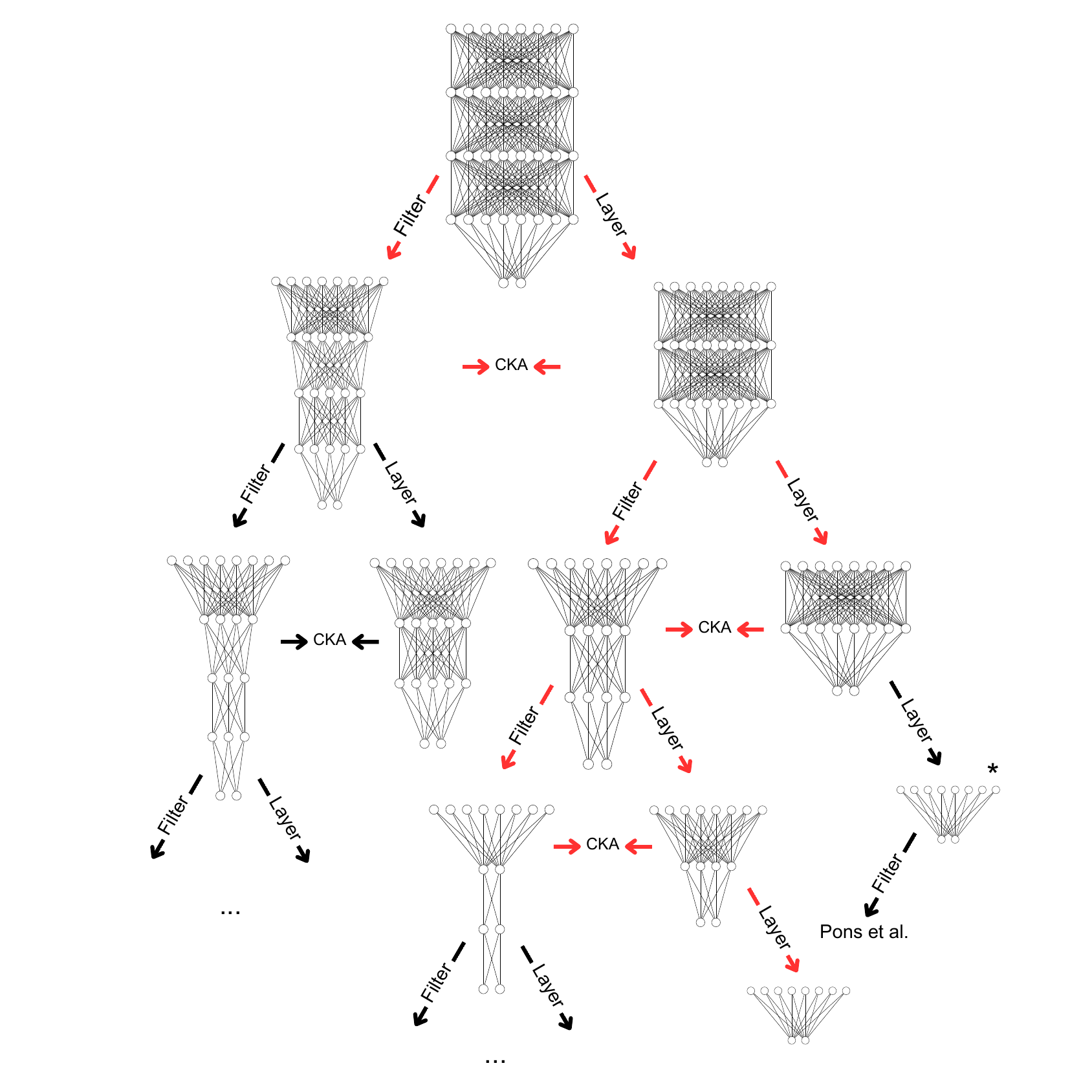}
\includegraphics[width=\columnwidth]{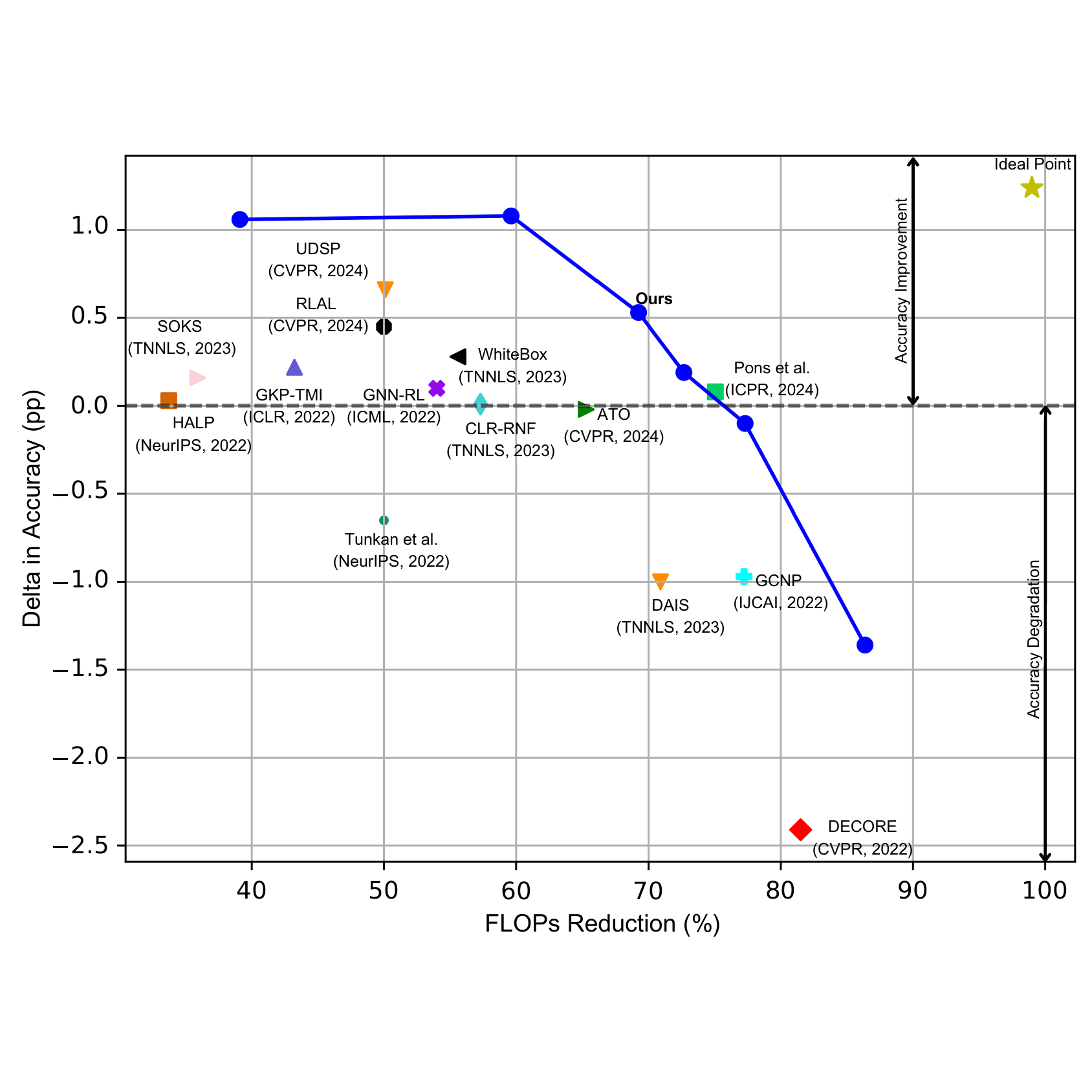}
\caption{\textbf{Left.} A binary decision tree of potential pruning outcomes and the paths (red arrows) to follow using our approach. The process starts with an unpruned deep network (parent), for which we generate two subnetworks (pruned): one from layer pruning and the other from filter pruning. Then, with a similarity metric (i.e.,  Centred Kernel Alignment -- CKA), we evaluate the resulting subnetworks to determine and select the one that better preserves the representation of its parent. Our method iteratively repeats this strategy until no more structures (filters or layers) remain for removal or until reaching a user-defined number of iterations. To accomplish this, the subnetwork with the highest similarity to its parent (i.e., the selected subnetwork candidate) becomes the parent for the next iteration (the next depth level in the tree). The example illustrates our method over four pruning iterations, where it systematically eliminated structures in the following sequence: layers, filters, layers, and layers. \textbf{Right.} Comparison with state-of-the-art techniques on the standard ResNet56 + CIFAR-10 setting. Our method significantly outperforms existing layer and filter pruning techniques by a large margin. Specifically, compared to other pruning approaches, our method removes over 80\% of FLOPs (Floating Point Operations) with negligible delta in accuracy. In contrast, other methods degrade accuracy when operating at such high levels of FLOPs reduction. In summary, our method achieves better trade-offs in accuracy and computational efficiency (measured by FLOPs). These results confirm how our approach enables better pruning by making informed choices between layers and filters, surpassing most state-of-the-art techniques. (For illustration purposes only, we intentionally misuse notation and indicate the ideal point on the top right.)
}

\label{fig:teaser}
\end{figure*}

Efforts towards compression of deep models confirm the effectiveness of pruning methods in reducing computational demands without degrading the predictive ability of models~\cite{Cheng:2024}. Most pruning methods remove small structures composing a model such as weights or filters~\cite{Cheng:2024, He:2023}. The first, a.k.a unstructured pruning, obtains higher compression rates but requires specialized hardware for sparse computing. The latter, namely structured pruning, offers hardware-agnostic benefits and competitive sparsity rates. In contrast to these popular forms of pruning, a parallel line of study suggests that removing layers achieves both superior predictive preservation and better enhances depth-related performance metrics such as latency~\cite{Pons:2024,Kim:2024}. However, layer pruning is not without its drawbacks. For example, due to technical details, the number of layers available for removal is notably smaller than filters; thus, limiting the maximum compression level this family of pruning can obtain (roughly $75.5\%$ of Floating Point Operations -- FLOPs -- on the popular CIFAR10 + ResNet56 setting). Figure~\ref{fig:teaser} (\textbf{Left}) illustrates this behavior, where a typical layer-pruning method consistently removes layers until none remain among those eligible for removing. This means they take the path to the right until reaching a leaf (see symbol * in Figure~\ref{fig:teaser} (\textbf{Left})). As a concrete example, Pons et al.~\cite{Pons:2024} eliminate layers until no more layers are available for pruning. After this point, to achieve even higher compression rates, their method starts removing neurons (i.e., filters), following the path to the left from the leaf (indicated by *) in Figure~\ref{fig:teaser} (\textbf{Left}). Technically speaking, existing filter-pruning approaches share similar limitations. Considering that the literature focuses on eliminating individual structures~\cite{He:2023,Cheng:2024}, filters or layers but not both, a naive strategy for obtaining high-level compression is to eliminate different structures throughout a pruning process.

From the previous discussion, a possible approach to overcome the limitations of existing structured pruning involves alternating between structures. In this direction, a natural question arises: \emph{how do we decide which structure to remove during pruning?} Figure~\ref{fig:teaser} (\textbf{Left}) allows restructuring the question as follows: \emph{given a network (the parent\footnote{Following Manson et al.~\cite{Manson:2024}, we use the term parent referring to a network that generates the subnetworks (i.e., pruned models).}), should we take the right path (removing layers) or the left path (removing filters) as pruning progresses?}

To accomplish this, we propose systematically deciding between eliminating layers or filters as pruning progresses. An overview of our method is the following. From a network, we generate two candidate subnetworks: one through layer and filter pruning, respectively. Afterwards, we use Centred Kernel Alignment (CKA)~\cite{Kornblith:2019} to compare the internal representation of the candidates with the network that generated them (namely, parent). Then, we take the candidate with higher similarity. From the perspective of Figure~\ref{fig:teaser} (\textbf{Left}), we use CKA to guide our choice between going to the left (eliminate filters) or right (eliminate layers) path for a given network.
The rationality behind this process involves evaluating how well a candidate subnetwork preserves the representation of its parent: greater similarity indicates a proper candidate.

The process above comprehends one iteration of our method. Following previous works~\cite{Pons:2024,Muralidharan:2024,Shen:2022,Ganjdanesh:2024}, we perform this process iteratively and, for each iteration, our method chooses the candidate greedily. This means it selects the locally proper candidate at each pruning iteration (one level in Figure~\ref{fig:teaser} (\textbf{Left})) and ensures computational efficiency with $O(log (n))$ complexity, where $n$ is the set of all possible candidate subnetworks.

\noindent
\textbf{Research Statement and Contributions.} Overall, our work has the following research statement: \emph{Given two candidate subnetworks (pruned models), one from layer pruning and the other from filter pruning, we can effectively decide which to choose by selecting the one with the highest representation similarity to its parent. Iteratively and greedily repeating this process provides highly sparse models that preserve the predictive ability of the overparameterized original  model (unpruned).} We confirm this statement across a broad range of experimental settings, datasets, and architectures.

Among our contributions, we highlight the following. We introduce a new form of pruning that eliminates both filter and layer from deep models and leverages the best of existing techniques. From a practical perspective, our method contributes to achieving a new milestone of computational gains. The proposed method is not confined to removing only one type of structure and integrates the benefits of both filter and layer methods. From a theoretical perspective, we demonstrate that preserving representation similarity between the parent model and a candidate subnetwork (from layer and filter pruning) indicates an effective strategy for systematically deciding which structure to eliminate. This enables maintaining predictive ability during pruning, even at significantly high compression rates.

Figure~\ref{fig:teaser} (\textbf{Right}) supports the previous statements by introducing our results alongside other state-of-the-art methods. Specifically, we achieve a $72.67\%$ reduction in FLOPs without degrading accuracy and reach a milestone of $86.37\%$ with a negligible drop.

Extensive experiments across a broad range of benchmarks and architectures confirm our method outperforms state-of-the-art pruning techniques. Specifically, we surpass existing methods by achieving $61.99\%$ and $71.31\%$ FLOPs reduction on ResNet32 and ResNet44 while increasing accuracy. On deeper architectures, we obtain a milestone of $86.37\%$ and $95.82\%$ FLOPs reduction on ResNet56 and ResNet110. In terms of GreenAI~\cite{Lacoste:2019,Faiz:2024,anonymous2025holistically}, these results comprehend an important step toward reducing carbon emissions linked to energy consumption during model deployment. Our pruned models also reduce financial costs associated with training and fine-tuning 
by up to $83\%$.

With the popularization of (large) foundation models~\cite{Dubey:2024,Block:2025} and the increasing need for efficient models, we believe our results open a new chapter for achieving higher levels of compression by breaking down the barriers of existing strategies that rely solely on one structure.

%% file: Sections/RelatedWork.tex
\section{Related Work}\label{sec:related}
Recent literature categorizes this family of methods into two approaches~\cite{He:2023,Cheng:2024}: structured and unstructured. Structured pruning removes structural components of the neural network, such as neurons (filters, attention heads) and layers. Overall, these methods assign a score to each structure based on specific criterion and remove structures accordingly. In contrast, unstructured pruning focuses on individual weights. Although this form of pruning achieves higher sparsity levels, deploying it in real-world scenarios demands specific hardware support, whereas structured pruning avoids such constraints. Additionally, most pruning methods optimize for specific structures, such as filters or layers, being unable to leverage the possibility of removing more than one type of structure (i.e., orthogonality)~\cite{He:2023,Cheng:2024}. As a concrete example, Gao et al.~\cite{Gao:2024} proposed a bi-level optimization to integrate benefits of both efficiency and storage. To accomplish this, their method combines dynamic (input-dependent, adapting during inference) and static (fixed, not changing during inference) filter pruning. From the lens of layer pruning, Ganjdanesh et al.~\cite{Ganjdanesh:2024} addressed the problem with reinforcement learning (RL), pruning iteratively while updating both RL and model parameters. Wu et al.~\cite{Wu:2024} proposed a method for pruning while working alongside a controller network to avoid falling into local optima. Kim et al.~\cite{Kim:2024} compressed models by using a layer merge method that removes layers and activation functions. Yang et al.~\cite{Yang:2024} introduced a method that collapses later layers into earlier ones, reducing model size while preserving its structure. In a similar direction, Gromov et al.~\cite{Gromov:2025} focused on analyzing where knowledge is stored: in deeper or shallower layers.

Despite positive results, the aforementioned works focus on pruning only one type of structure. Therefore, removing both structures simultaneously remains an open problem. In contrast, our method fills this gap. It is worth mentioning that the work by Pons et al.~\cite{Pons:2024} constitutes a preliminary effort in this direction. Specifically, the authors introduced a layer-pruning technique. Unfortunately, the number of layers constrains the maximum computational benefit the method can achieve (other layer-pruning methods suffer from the same limitation). To mitigate this problem, their method leverages the orthogonal nature of filter and layer pruning and begins removing filters only after eliminating all possible layers. However, the authors observed that this approach fails to preserve generalization, highlighting the complexity of synergizing filter and layer pruning. This supports the core idea behind our method as it carefully and systematically chooses among the type of structure.

Closely related to ours, Muralidharan et al.~\cite{Muralidharan:2024} discover subnetworks by alternating between width (i.e., filters and attention heads) and depth (i.e., layers) pruning. Unfortunately, their process relies on an empirical trial-and-error mechanism. The challenge with this strategy lies in understanding the process behind deriving these rules and determining their applicability to all models. In contrast, our method follows a systematic and empirical-agnostic approach by employing similarity metrics to select subnetworks from filter or layer pruning automatically. Therefore, we believe our work opens a new chapter in pruning by exploring structural possibilities as decisions, thereby overcoming the inherent limitations of structure-dependent pruning.

%% file: Sections/Methodology.tex
\section{Preliminaries and Proposed Method}
\noindent
\textbf{Problem Statement.} 
According to recent literature~\cite{He:2023,Cheng:2024,kim2024shortened}, layer and filter strategies offer diverse and complementary advantages in reducing the computational cost of deep models. For example, filter pruning enables a high compression rate while layer pruning promotes notable inference improvements. Therefore, to leverage the best of both worlds, we define the following problem statement: \emph{how to identify the most promising structure to remove across an iterative pruning process?} Solving this problem enables obtaining shallower (layer pruning) and narrower (filter pruning) models after repeating the process multiple times.

The above formulation naturally suggests a naive solution: if we aim to remove layers or filters during a pruning process, why not just choose randomly which one to eliminate?
Previous studies endorse this naive solution~\cite{Li:2022:CVPR}, confirming the potential of random pruning. We demonstrate that systematically and carefully selecting among layers or filters (as explained below) leads to better results for this problem.

\noindent
\textbf{Definitions.} Let $\mathcal{F}$ denote a neural network trained on a dataset $\mathbf{X} = \{x_i\}^{D}_{i=1}$ with corresponding labels $\mathbf{Y} = \{y_i\}^{D}_{i=1}$, where $D$ is the size of the dataset.
Consider a pruning criterion $c$ that assigns a score (i.e., importance) to each structure (filter or layer) composing the network. Based on these scores, a pruning algorithm eliminates the less important structures.

Applying $c$ to a neural network to prune filters results in a subnetwork $\mathcal{F}^{'}_{f}$. Similarly, using it to prune layers produces a subnetwork $\mathcal{F}^{'}_{l}$. As we shall see, the core of our method relies on comparing these pruned networks with their parent. To accomplish this, consider a similarity metric that takes the internal representations (a.k.a feature maps) of two neural networks as input and provides a measure of their similarity. In this work, we employ the Centered Kernel Alignment (CKA) similarity metric~\cite{Kornblith:2019}, defined as the normalized Hilbert-Schmidt Independence Criterion (HSIC) by Gretton et al.~\cite{Gretton:2005}. Formally, let $k$ and $l$ be two kernels, the empirical estimator of HSIC with $m$ examples for the internal representation extraction is:

\begin{equation}
	\label{eq:hsic}
	HSIC(K, L) = \frac{1}{(m-1)^2}tr(KHLH),
\end{equation} where $K_{ij} = k(x_i, x_j)$, $L_{ij} = l(y_i, y_j)$ and $H_{ij} = \delta_{ij} - m^{-1}$. Here, $\delta_{ij} = 1$ if $i = j$, and $\delta_{ij} = 0$ otherwise.

Let $R$ be the internal representation from the parent model $\mathcal{F}$ and $R_{\mathcal{F}^{'}}$ the internal representation from a pruned model $\mathcal{F}^{'}$. Following Kornblith et al. \cite{Kornblith:2019}, we compute the CKA in terms of
{
\small
\begin{equation}
	CKA(R,R_{\mathcal{F}^{'}}) = \frac{HSIC(R,R_{\mathcal{F}^{'}})}{\sqrt{HSIC(R,R) \times HSIC(R_{\mathcal{F}^{'}},R_{\mathcal{F}^{'}})}}.
	\label{eq:cka}
\end{equation}
}

The output from $CKA(R,R_{\mathcal{F}^{'}}) \in [0,1]$, where higher values indicate greater similarity between the representations. For simplicity, we slightly abuse notation and apply CKA similarity using two networks directly as inputs -- the parent model and its pruned version. In practice, however, CKA compares two matrices that characterize the internal representation (i.e., feature maps) of the models.

\noindent
\textbf{Proposed Method.} 
Building upon the previous definition, our method operates as follows. First, it produces a layer $\mathcal{F}^{'}_{l}$ and filter $\mathcal{F}^{'}_{f}$ pruned version (subnetworks) from a network $\mathcal{F}$. At this step, we follow previous work and generate subnetworks with similar capacity~\cite{Jordao:2023}, ensuring a consistent comparison among them. Additionally, to ensure a fair comparison, we generate the subnetworks using the same pruning criterion $c$.
An important point regarding $c$ arises here: an effective criterion for removing filters may be inadequate when applied directly to remove layers~\cite{Jordao:2023,Pons:2024}. To deal with this issue, we apply the KL-divergence criterion by Luo et al.~\cite{Luo:2020}. This criterion measures Kullback-Leibler divergence between the softmax distribution of the unpruned network and the network without a potential component. Therefore, it applies to any structure (weight, filter or layer) and is unaffected by common pitfalls of traditional filter criteria (e.g., $\ell_1$-norm) when applied to layers~\cite{Pons:2024}. Interestingly, in Appendix~\ref{sec:ablation}, we evaluate our method using the popular $\ell_1$-norm criterion~\cite{He:2023,Cheng:2024} and confirm it is effective regardless of the criterion.

After the above steps, our method considers $\mathcal{F}^{'}_{l}$ and $\mathcal{F}^{'}_{f}$ as pruning candidates and must select one of them. In other words, these candidates form a branch in a binary decision tree (Figure~\ref{fig:teaser} (\textbf{Left})), and the method must choose a path. To accomplish this, the central idea of our method is to select the one with the highest similarity w.r.t $\mathcal{F}$ (with the network that produced them -- parent); thereby preserving predictive capability of the original neural network $\mathcal{F}$ at the root of the tree (Figure~\ref{fig:teaser} (top)).
To guide this decision, we apply the CKA similarity metric (Equation~\ref{eq:cka}) and formulate the expression:
\begin{equation}
	\mathcal{F} \leftarrow 
	\begin{cases}
		\mathcal{F}^{'}_l    &   \text{if } CKA(\mathcal{F}, \mathcal{F}^{'}_l) \geq CKA(\mathcal{F}, \mathcal{F}^{'}_f) \\
		\mathcal{F}^{'}_f    &   \text{otherwise. }      
	\end{cases}
	\label{eq:selectpruning}
\end{equation}
Due to the iterative nature of our method (details below), in Equation~\ref{eq:selectpruning}, we slightly abuse notation\footnote{Another way to express this formulation is $\mathcal{F}^k \leftarrow (\cdot)$, where $k$ indicates the $kth$ iteration.} and assign $\mathcal{F}$ to its pruned version for easy of exposition. Finally, we apply a standard (i.e., supervised) fine-tuning, as this step helps recover predictive ability after pruning~\cite{Manson:2024,kim2024shortened,Muralidharan:2024}. Note that when the similarity of both subnetworks is equal, our method prefers to remove the layer, as this form of pruning promotes additional performance gains~\cite{kim2024shortened,Pons:2024}. In this sense, Equation~\ref{eq:selectpruning} allows us to easily incorporate inference-aware thought layer pruning by simply penalizing the left part of with a constant $\epsilon$ (i.e.,  $CKA(\mathcal{F}, \mathcal{F}^{'}_f) + \epsilon$).
\begin{algorithm}[!t]
	\small
	\caption{Proposed Method}
	\label{alg::pruning}
	\begin{algorithmic}[1]
		\item[] \textbf{Input:} Trained Neural Network $\mathcal{F}$, Pruning Criterion $c$, Number of Iterations $K$, Training samples $\textbf{X}$ and the respective labels $\textbf{Y}$
		\item[] \textbf{Output:} Pruned version of $\mathcal{F}$\\
		\FOR{$k \leftarrow 1$ {\bfseries to} $K$}\label{it_line}
		\STATE $\mathcal{F}^{'}_{l} \leftarrow \mathcal{F} \setminus c(\mathcal{F}) $ $\triangleright$ Remove unimportant layers based on $c$\\
		\STATE$\mathcal{F}^{'}_{f} \leftarrow \mathcal{F} \setminus c(\mathcal{F}) $ $\triangleright$ Remove unimportant filters based on $c$\\
		\IF{ $CKA(\mathcal{F}, \mathcal{F}^{'}_{l})$ $\geq$ $CKA(\mathcal{F}, \mathcal{F}^{'}_{f})$}\label{alg::cka_comparison}
		\STATE $\mathcal{F} \leftarrow \mathcal{F}^{'}_{l}$  $\triangleright$ $\mathcal{F}$ becomes its layer pruned version\
		\ELSE
		\STATE $\mathcal{F} \leftarrow \mathcal{F}^{'}_{f}$  $\triangleright$ $\mathcal{F}$ becomes its filter pruned version\
		\ENDIF
		\STATE Update $\mathcal{F}$ via standard fine-tuning on  $\textbf{X}$ and $\textbf{Y}$\label{alg::ft}
		\ENDFOR	
	\end{algorithmic}
\end{algorithm}

Algorithm~\ref{alg::pruning} summarizes the steps composing our method. From Algorithm~\ref{alg::pruning}, we highlight two key observations. (i) Line 4 indicates a branch (i.e., decision) in Figure~\ref{fig:teaser} (\textbf{Left}). (ii) The input to the next ($k+1$) pruning iteration is the model pruned in the previous ($k$) iteration, see lines $5$ and $7$. Importantly, this is what ensures the algorithm produces shallower and narrower models after multiple iterations ($K$). 

Regarding the time-complexity, since for each iteration ($k$) we reduce the problem in terms of $\frac{n}{2^{K}}$, where $n$ is the number of nodes in the tree, our method has $O(log(n))$ time complexity. In this direction, we iterate using a sufficiently large $K$; thus, obtaining notably shallow and narrow models.

%% file: Sections/Experiments.tex
\section{Experiments}\label{sec:experiments}
\noindent
\textbf{Experimental Setup.} We conduct the experiments on CIFAR-10 and ImageNet using different ResNet architectures~\cite{He:2016}. These settings are a standard choice to assess the effectiveness of pruning \cite{Wu:2024,Ganjdanesh:2024,Gao:2024,He:2023,Pons:2024}. 

Following previous works~\cite{Muralidharan:2024,Manson:2024,Sun:2024} before comparing the two pruned subnetworks in Algorithm~\ref{alg::pruning} (line~\ref{alg::cka_comparison}), we also perform a 10-epoch fine-tuning. We omit this step from Algorithm~\ref{alg::pruning} to maintain the simplicity and clarity of our approach.

Overall, we keep the experimental setup as simple as possible to highlight the potential and advantages of our method compared to more elaborate setups on pruning such as knowledge distillation~\cite{Chen:2019,Zhou:2022,Muralidharan:2024}. We also conduct experiments on Transformers and detail our findings in Appendix~\ref{sec:transformer}.

To summarize the results, we adopt common approaches and present the delta in accuracy in percentage points (pp)~\cite{He:2023}. As shown in Figure~\ref{fig:teaser} (\textbf{Right}), a negative delta indicates a degradation in accuracy, while a positive delta reveals an improvement. Finally, for each division of FLOP reduction ($\%$), we highlight the best results in bold in terms of delta in accuracy and FLOP reduction, separately.

\begin{table}[!b]
	\footnotesize
	\addtolength{\tabcolsep}{-3pt}
	\renewcommand{\arraystretch}{1.1}
	\caption{Comparison with state-of-the-art methods on CIFAR-10.}
	\label{tab:main_resultsResNet56}
	\begin{tabular}{clcc}
		\hline
		& \multicolumn{1}{c}{Method} & $\Delta$ Acc.     & FLOPs (\%)     \\ \hline
		\multicolumn{1}{l|}{} & DECORE~\cite{Alwani:2022} (CVPR, 2022)    & (+) 0.08         & 26.30      \\
		\multicolumn{1}{l|}{}  		  &		HALP~\cite{Shen:2022} (NeurIPS, 2022)	 & (+) 0.03		& 33.72 \\
		\multicolumn{1}{l|}{}         &                   SOKS~\cite{Liu:2023} (TNNLS, 2023)       & (+) 0.16         & 35.91\\
		\multicolumn{1}{l|}{}         &                  Pons et al.~\cite{Pons:2024} (ICPR, 2024)          & \textbf{(+) 1.25}         & 37.52            \\ 
		\multicolumn{1}{l|}{}         &                  \textbf{Ours}           & (+) 1.06         & \textbf{39.10}           \\ \cline{2-4} 
		\multicolumn{1}{l|}{}         &                 GKP-TMI~\cite{Zhong:2022} (ICLR, 2022)     & (+) 0.22         & 43.23 \\
		\multicolumn{1}{l|}{}         &                 GCNP~\cite{Jiang:2022} (IJCAI, 2022)      & (+) 0.13         & 48.31            \\
		\multicolumn{1}{l|}{}         &                 Pons et al.~\cite{Pons:2024} (ICPR, 2024)          & (+) 0.86         & 48.78            \\ 
		\multicolumn{1}{l|}{}         &                 Tunkan et al.~\cite{Tukan:2022} (NeurIPS, 2022)       & (-) 0.65         & 50.00          \\
		\multicolumn{1}{l|}{}         &                 RLAL~\cite{Ganjdanesh:2024} (CVPR, 2024)       & (+) 0.45         & 50.00          \\
		\multicolumn{1}{l|}{}         &                 UDSP~\cite{Gao:2024} (CVPR, 2024)       & (+) 0.66         & 50.1          \\
		\multicolumn{1}{l|}{}         &                 \textbf{Ours}           & \textbf{(+) 0.95}         & \textbf{50.36}         \\ \cline{2-4}
		\multicolumn{1}{l|}{}         &                 \textbf{Ours}           & \textbf{(+) 0.91}         & 53.70            \\ 
		\multicolumn{1}{l|}{\multirow{2}{*}{ResNet56}}         &                 GNN-RL~\cite{Yu:2022}  (ICML, 2022)    & (+) 0.10         & 54.00            \\
		\multicolumn{1}{l|}{}         &                 ATO~\cite{Wu:2024} (CVPR, 2024)     & (+) 0.24         & 55.00  \\
		\multicolumn{1}{l|}{} 		&		WhiteBox~\cite{WhiteBox:2023} (TNNLS, 2023) & (+) 0.28         & \textbf{55.60}      \\ \cline{2-4}
		\multicolumn{1}{l|}{} 		&		CLR-RNF~\cite{Lin:2023} (TNNLS, 2023)    & (+) 0.01        & 57.30      \\
		\multicolumn{1}{l|}{} 		&		Pons et al.~\cite{Pons:2024} (ICPR, 2024)          & (+) 0.78        & 60.04      \\ 
		\multicolumn{1}{l|}{}         &                 \textbf{Ours}           & \textbf{(+) 0.96}         & \textbf{62.79}          \\ \cline{2-4}
		\multicolumn{1}{l|}{}         &                 ATO~\cite{Wu:2024} (CVPR, 2024)     & (-) 0.02         & 65.30  \\
		\multicolumn{1}{l|}{}         &                 \textbf{Ours}           & \textbf{(+) 0.53}         & \textbf{69.24}            \\ \cline{2-4}
		\multicolumn{1}{l|}{} 		&		DAIS~\cite{Guan:2023} (TNNLS, 2023)      & (-) 1.00         & 70.90      \\
		\multicolumn{1}{l|}{}		&		\textbf{Ours}           & \textbf{(+) 0.19}         & 72.67      \\
		\multicolumn{1}{l|}{}		&		Pons et al.~\cite{Pons:2024} (ICPR, 2024)          & (+) 0.08         & 75.05      \\
		\multicolumn{1}{l|}{}		&		GCNP~\cite{Jiang:2022} (IJCAI, 2022)    & (-) 0.97         & 77.22      \\
		\multicolumn{1}{l|}{}		&		\textbf{Ours}           & (-) 0.49         & \textbf{79.88}      \\ \cline{2-4}
		\multicolumn{1}{l|}{}		&		DECORE~\cite{Alwani:2022} (CVPR, 2022)     & (-) 2.41         & 81.50      \\
		\multicolumn{1}{l|}{}		&		\textbf{Ours}      & \textbf{(-) 1.36}         & \textbf{86.37} \\ \hline
		
		\multicolumn{1}{c|}{} & DECORE~\cite{Alwani:2022} (CVPR, 2022)	&(+) 0.38			&35.43			\\
		\multicolumn{1}{c|}{}       &                   GKP-TMI~\cite{Zhong:2022} (ICLR, 2022)	&(+) 0.64			& 43.31			\\
		\multicolumn{1}{c|}{}       &                     Pons et al.~\cite{Pons:2024} (ICPR, 2024) 					& (+) 1.37          & 44.73      
		\\ 
		\multicolumn{1}{l|}{}		&		\textbf{Ours}           & \textbf{(+) 1.47}         & \textbf{48.29}      \\ \cline{2-4}
		\multicolumn{1}{l|}{}		&		\textbf{Ours}           & \textbf{(+) 0.73}         & 59.95      \\ 
		\multicolumn{1}{c|}{}		&		DAIS~\cite{Guan:2023} (TNNLS, 2023)     & (-) 0.60         & 60.00      \\
		\multicolumn{1}{c|}{}		&		DECORE~\cite{Alwani:2022} (CVPR, 2022)     & 0.00          & \textbf{61.78}      \\ \cline{2-4}
		\multicolumn{1}{c|}{}		&		Pons et al.~\cite{Pons:2024}	(ICPR, 2024)				    & \textbf{(+) 0.89}         & 65.23      \\ 
		\multicolumn{1}{c|}{}		&		CRL-RNF~\cite{Lin:2023} (TNNLS, 2023)       & (+) 0.14          & 66.00      \\
		\multicolumn{1}{c|}{\multirow{1}{*}{ResNet110}}		&		WhiteBox~\cite{WhiteBox:2023} (TNNLS, 2023)  & (+) 0.62          & 66.00      \\
		\multicolumn{1}{c|}{}		&		Pons et al.~\cite{Pons:2024} (ICPR, 2024)				& (+) 0.80          & 67.10      \\
		\multicolumn{1}{l|}{}		&		\textbf{Ours}           & (+) 0.86         & \textbf{69.90}    \\ \cline{2-4}
		\multicolumn{1}{c|}{}		&		Pons et al.~\cite{Pons:2024} (ICPR, 2024)				& (+) 0.59          & 70.83      \\ 
		
		\multicolumn{1}{l|}{}		&		\textbf{Ours}           & \textbf{(+) 0.60}         & \textbf{73.16}      \\ \cline{2-4}
		\multicolumn{1}{l|}{}		&		\textbf{Ours}           & \textbf{(+) 0.49}         & 75.55      \\
		\multicolumn{1}{c|}{}		&		DECORE~\cite{Alwani:2022} (CVPR, 2022)       & (-) 0.79          &76.92      \\
		\multicolumn{1}{c|}{}		&		Pons et al.~\cite{Pons:2024} (ICPR, 2024)				& (+) 0.23         & \textbf{76.42}      \\ \cline{2-4}
		\multicolumn{1}{c|}{}		&		Pons et al.~\cite{Pons:2024} (ICPR, 2024)				& \textbf{(-) 0.41}         & 87.61      \\ 
		\multicolumn{1}{l|}{}		&		\textbf{Ours}           & (-) 2.91         & \textbf{95.82}      \\\hline
	\end{tabular}
\end{table}

\noindent
\textbf{Comparison with State-of-The-Art.} We start our analysis by comparing our method with modern and top-performance pruning techniques. Table~\ref{tab:main_resultsResNet56} summarizes the results on the standard FLOPs reduction and delta in accuracy compromises.

According to Table~\ref{tab:main_resultsResNet56}, our approach surpasses most existing techniques, regardless of whether it is pruned by layers or filters. On ResNet56, our best result (without degrading generalization) reduces $72.67\%$ of FLOPs while improving accuracy in $0.19$ pp. In terms of maintaining positive accuracy while achieving high compression, our method only underperform Pons et al.~\cite{Pons:2024}. Notably, their method is confined to not exceeding this value, as there are no more layers to remove (see the symbol * in Figure~\ref{fig:teaser} (\textbf{Left})). In particular, all methods suffer from this constraint since they are one structure pruning. It turns out that while state-of-the-art techniques opt to remove only layers (path to the right) or only filters (path to the left), they overlook the middle ground. Focusing on only one type of structure ignores the fact that there is much more to explore by systematically alternating between them. The results in Table~\ref{tab:main_resultsResNet56} confirm the superiority of making informed pruning decisions. When considering only FLOP reductions, our approach outperforms all other works, achieving a reduction of $86.57\%$.

On ResNet110, the previous behavior repeats itself, and considering the level of compression, our approach achieves a FLOP reduction of $95.82\%$ while exhibiting a negligible delta in accuracy. To the best of our knowledge, these values represent the highest reduction achieved solely through pruning, and recent surveys confirm this~\cite{He:2023,Cheng:2024}. Finally, Table~\ref{tab:main_resultsImageNet} introduces our results on ImageNet using ResNet50, where we achieve performance on par with or surpassing the state-of-the-art across different ranges of FLOP reduction.
\begin{table}[!b]
	\footnotesize
	\addtolength{\tabcolsep}{-3pt}
	\renewcommand{\arraystretch}{1.1}
	\caption{Comparison with state-of-the-art pruning methods on ImageNet using ResNet50. $\Delta$ Acc. considers Top1 accuracy.}
	\label{tab:main_resultsImageNet}
	\begin{tabular}{clcc}
		\hline
		& \multicolumn{1}{c}{Method} & $\Delta$ Acc.     & FLOPs (\%)     \\ \hline
		\multicolumn{1}{l|}{} &	DECORE~\cite{Alwani:2022} (CVPR, 2022)      & (+) 0.16             & 13.45          \\
		\multicolumn{1}{l|}{}  		  &	\textbf{Ours}   & \textbf{(+) 2.09}             & \textbf{16.98}          \\ \cline{2-4}
		\multicolumn{1}{l|}{}  		  &	SOSP~\cite{Nonnemacher:2022} (ICLR, 2022)   & (+) 0.41             & 21.00          \\
		\multicolumn{1}{l|}{}  		  &	GKP-TMI~\cite{Zhong:2022} (ICLR, 2022)      & (-) 0.19             & 22.50          \\
		\multicolumn{1}{l|}{}  		  &	Pons et al.~\cite{Pons:2024} (ICPR, 2024)                                & (+) 1.11             & 22.64          \\
		\multicolumn{1}{l|}{}  		  &	\textbf{Ours}   & \textbf{(+) 1.61}             & \textbf{22.64}          \\ \cline{2-4}
		\multicolumn{1}{l|}{}  		  &	SOSP~\cite{Nonnemacher:2022} (ICLR, 2022)   & (+) 0.45             & 28.00          \\
		\multicolumn{1}{l|}{\multirow{2}{*}{ResNet50}}  		  &	Pons et al.~\cite{Pons:2024} (ICPR, 2024)                                  & (+) 0.74             & 28.30          \\ 
		\multicolumn{1}{l|}{}  		  &	\textbf{Ours}   & \textbf{(+) 1.07}             & \textbf{28.30}          \\ \cline{2-4}
		\multicolumn{1}{l|}{}  		  &	CLR-RNF~\cite{Lin:2023} (TNNLS, 2023)       & (-) 1.16             & 40.39          \\
		\multicolumn{1}{l|}{}  		  &	DECORE~\cite{Alwani:2022} (CVPR, 2022)      & (-) 1.57             & 42.30          \\
		\multicolumn{1}{l|}{}  		  &	SOSP~\cite{Nonnemacher:2022} (ICLR, 2022)   & (-) 0.94             & 45.00          \\
		\multicolumn{1}{l|}{}  		  &	\textbf{Ours}   & 	(-) 0.98             & 45.28          \\
		\multicolumn{1}{l|}{}  		  &	Pons et al.~\cite{Pons:2024} (ICPR, 2024)  & (-) 0.90  & 45.28          \\ \multicolumn{1}{l|}{}  		  &	WhiteBox~\cite{WhiteBox:2023} (TNNLS, 2023) & \textbf{(-) 0.83}             & \textbf{45.60}          \\
		\hline
	\end{tabular}
\end{table}

The key factor behind the aforementioned breakthroughs is the careful selection of which structures to remove during the pruning process. Manson et al.~\cite{Manson:2024} argue that knowing which parameters to remove is more important than quantity, reinforcing the argument behind our method, although it explores a structural perspective. Besides, the concept of pruning seeks to maintain generalization while enhancing compression, resulting in a shallower and/or narrower model. By using CKA as a similarity metric, we focus directly on the internal representation of the neural network. When we adopt it as a selection criterion, we retain the best subnetwork that represents its parent in terms of internal representation, allowing continued pruning while preserving accuracy.

Following by our findings, the previous results highlight the potential of employing our systematic approach that alternates between layers and filters, rather than solely pruning one type of structure. Moreover, we confirm our research statement: the similarity between the subnetwork and its parent allows to obtain highly sparse models that preserve the predictive ability of the overparameterized original (unpruned) model.

We now turn our attention to analyzing the effectiveness and robustness of the proposed method in different settings.

\noindent
\textbf{Effectiveness in Shallow Architectures.} 
One of the keys to successful pruning lies in the overparameterized nature of networks, particularly in deep models. 
The previous experiments confirm the effectiveness of our method on this family of models. In this experiment, we highlight how the proposed method operates well to shallow architectures. For this purpose, we use the shallow versions of ResNet: ResNet32/44.

By applying our method on ResNet32 and ResNet44, we observe the same behavior as in their deeper versions (ResNet56/110). More concretely, we reduce $61.99\%$ and $71.31\%$ of the FLOPs on ResNet32 and ResNet44, respectively, without compromising accuracy. Interestingly, to obtain more than $95\%$ FLOP reduction on ResNet32, our method exhibited a notable drop in accuracy of $7.63$ pp. We highlight that this is the only instance in which we observe a severe drop through our experiments, yet no previous works report such an achievement on this architecture.

Similarly to the deeper models, we also compare these results with state-of-the-art methods for the smaller architectures. Our approach accomplishes notable results, outperforming existing methods.
Our approach pushes the boundaries of compression, reducing $87.06\%$ and $83.19\%$ of the FLOPs on ResNet32/44, with only a small drop in accuracy (e.g., $2.83$ pp and $1.26$ pp). We provide detailed results in Appendix~\ref{sec:shallow}.

Overall, the previous discussion confirms that our method performs effectively on both deep and shallow networks and establishes compatibility with modern architectures, as shown in Appendix~\ref{sec:transformer} with experiments on Transformers.

\noindent
\textbf{Is CKA Better than Random Choice?} 
As we mentioned in our problem definition, a naive solution for selecting between eliminating filters and layers during an iterative pruning process is simply to choose randomly (i.e., a random selection). In this experiment, we demonstrate that random selection is inefficient and leads to poorly pruned models. For this purpose, we prune the architectures ResNet32/44/56 by randomly selecting whether to remove filters or layers (namely, Random Walk). In other words, we replace line~\ref{alg::cka_comparison} in Algorithm~\ref{alg::pruning} with a coin flip. Importantly, we compare our method with Random Walk, varying only the selection mechanism, while all other settings (i.e., initialization, training/fine-tuning epochs, data augmentation and so on) remain identical. From these settings, we force pruning until the pruned model exhibits no degradation (positive delta) compared to the original, unpruned model. Due to the stochastic nature of the Random Walk, we ran it three times and averaged the results.

Table~\ref{tab:experiments} summarizes the results in terms delta in accuracy and FLOPs reduction. This table clearly shows that our method outperforms random selection across the architectures. Importantly, although the results appear similar in terms of accuracy, Random Walk stops pruning prematurely. Specifically, on ResNet32/56, after removing $55.83\%$ and $69.64\%$ of FLOPs, Random Walk starts degrading accuracy. In particular, when the models become deeper, it is unable to remove more than $70\%$ without degrading accuracy. Our method, on the other hand, successfully keeps removing more structures.

Overall, this experiment corroborates that carefully choosing the type of structure to eliminate, rather than selecting randomly, allows us to achieve significant FLOP reduction, as the pruning proceed further without compromising accuracy. In our initial analysis, we considered using some heuristics such as A* and branch-and-bound. However, their time complexities are computationally prohibitive due to the fine-tuning steps. Therefore, we do not extend experiments using these and other classical optimization mechanisms.

\begin{table}[!t]
	\footnotesize
	\centering
	\addtolength{\tabcolsep}{-3pt}
	\renewcommand{\arraystretch}{1.1}
	\caption{Comparison with random selection baseline (Random Walk) on Cifar-10 using different ResNets. Although Random Walk highlights positive accuracy, it stops pruning earlier.}
	\label{tab:experiments}
	\begin{tabular}{clcc}
		\hline
		& \multicolumn{1}{c}{Method} 					& $\Delta$ Acc.     & FLOPs (\%)     \\ \hline
		\multicolumn{1}{l|}{\multirow{2}{*}{ResNet32}}  		  					& Random Walk	 	& +0.12		& 55.83 \\
		\multicolumn{1}{l|}{}         					& \textbf{Ours }    & +0.09     & \textbf{61.99} \\ \hline
		
		\multicolumn{1}{l|}{\multirow{2}{*}{ResNet44}}  		  					& Random Walk	   & +0.21	& 69.74 \\
		\multicolumn{1}{l|}{}         					& \textbf{Ours}    & +0.66      & \textbf{71.31} \\  \hline
		
	\multicolumn{1}{l|}{\multirow{2}{*}{ResNet56}}  		  					& Random Walk	   & +0.19	& 69.64 \\
		\multicolumn{1}{l|}{}         					& \textbf{Ours}    & +0.19      & \textbf{72.67} \\  \hline
	\end{tabular}
\end{table}

\noindent
\textbf{Do High-capacity Models Favor Certain Structures?} 
In this experiment, we aim to determine if model capacity influences the preference for removing specific structures. Particularly, analyzing such a behavior is important to assess if our method is not converging to a trivial solution: remove one structure until complete and then start eliminating the other one. In order to verify this, we observe the initial pruning iterations. 

On shallow architectures, ResNet32/44, our method tends to remove more layer than filter. For example, with $10$ iterations of pruning (i.e., $K=10$ in Algorithm~\ref{alg::pruning}), we observe a ratio of layers and filters removed of $7$/$3$, on both architectures, where the first and second values indicate the number of occurrences our method eliminates layer and filters, respectively. 

Building upon the above analysis, we extend our observations to deeper models, including ResNet56 and ResNet110. For easy of exposition, we denote algorithm decisions as L (Layer) and F (Filter), and represent $i$ and $j$ consecutive layer and filter decisions as L$^{i}$ and F$^{j}$, respectively. In the initial iterations on these deep and high-capacity models, a clear pattern emerges. The pattern on ResNet56 does not persist for long, it alternates between filters and layers for $6$ iterations. On ResNet110, the pattern is more consistent, following the decision block (L, F$^{2}$) three times, then L$^{3}$, and repeating this sequence once more.	
	
In general, previous experiments reveal a transition in preference: \emph{the shallower the architecture, the greater the tendency to choose layers, while deeper models tend to prefer filters. But regardless of the architecture none structure is prevalent.}
	 
\noindent
\textbf{The Role of Fine-tuning.}
Throughout our analysis, we note that without fine-tuning, our method favors eliminating layers instead of filters. It turns out that the internal representations become inconsistent due to variations in the magnitude of the weights. Previous works corroborate this behavior~\cite{Manson:2024,Sun:2024,Muralidharan:2024}, where the authors argue that after pruning, subnetworks lose capacity. Fortunately, only a few epochs of fine-tuning are enough to restore subnetworks to an accuracy close to, or even higher than, their parent. Particularly, Manson el al.~\cite{Manson:2024} also highlight that if a pruned network does not recover within one epoch, it suggests the network is \emph{disconnected} from the loss manifold. For this reason, in both works~\cite{Manson:2024,Sun:2024}, the authors apply a $10$-epoch fine-tuning before continuing the process.

Building upon the prior discussion, performing fine-tuning before the comparison process is crucial for making a more informed choice in our approach. We note that after fine-tuning the method favors no structure; therefore, this step adheres to the principle of recovering the subnetwork and determining which one is more similar to its parent.

It is worth mentioning that as we follow this recovery strategy, there is no effect on our fine-tuning process in the Algorithm~\ref{alg::pruning} (line 9), as we discount the amount of $10$-epoch from the remaining fine-tuning epochs.

\noindent
\textbf{Invariance to Similarity Metrics.} This experiment shows that our method is robust to similarity metrics, achieving positive results across various similarity metrics. These metrics assess the proximity between objects, variables, models, and so forth. In this work, we leverage a similarity metric to identify the pruned subnetwork most similar to the parent, and although we use CKA, our algorithm supports other similarity metrics. To illustrate this, we conduct experiments using Linear~\cite{Williams:2021}, Gaussian Stochastic and Wasserstein~\cite{Duong:2023} metrics. According to previous works~\cite{Duong:2023,Williams:2021}, such metrics emerge as potential alternatives to CKA. Following our previous experiment, we indicate the algorithm decisions as L (Layer) and F (Filter) until the method achieve the subnetwork with highest FLOP reduction with a positive delta in the accuracy. 

Using the Linear metric, our method selected the path F$^{4}$, L, F, achieving a $56.18\%$ FLOP reduction. With the Gaussian Stochastic metric, our method pruned only layers with the path L$^{7}$, achieving $47.78\%$ of FLOP reduction. Finally, employing the Wasserstein metric it selected the path L, F, L$^{2}$, F, L$^{2}$, resulting in a $52.5\%$ FLOP reduction.

Table~\ref{tab:similarity} summarizes the results alongside with the result on ResNet32 with CKA. These results confirm that our method indicates robustness to different similarity metrics. Aiming to consistently choose the best structure to remove, we show that our method is similarity metric-agnostic, and we highlight that using CKA provides better outcomes. Importantly, CKA computation is more efficient than other metrics (due to space constraints, we refer to the works by Duong et al.~\cite{Duong:2023} and Williams et al.~\cite{Williams:2021} for additional information).

\begin{table}[!h]
	\footnotesize
	\centering
	\addtolength{\tabcolsep}{-3pt}
	\renewcommand{\arraystretch}{1.1}
	\caption{Comparison with other similarity metrics on Cifar-10. These results reinforce that our method is similarity metric-agnostic strategy.}
	\label{tab:similarity}
	\begin{tabular}{clcc}
		\hline
		& \multicolumn{1}{c}{Method} 					& $\Delta$ Acc.     	& FLOPs (\%)     \\ \hline
		\multicolumn{1}{l|}{}  		  					& Linear		 		& +0.24		& 56.18 \\
		\multicolumn{1}{l|}{\multirow{2}{*}{ResNet32}} 	& Gaussian Stochastic 	& +0.71     & 47.78      \\
		\multicolumn{1}{l|}{}  		  					& Wasserstein	 		& +0.32		& 52.50 \\
		\multicolumn{1}{l|}{}         					& \textbf{Ours (CKA)}   & +0.09     & 61.99 \\ \hline
	\end{tabular}
\end{table}

\noindent
\textbf{Robustness to Adversarial Samples.} As deep models assume greater significance in critical scenarios such as medical diagnosis or financial systems, robustness to out-of-distribution (OOD) and adversarial samples becomes equally important as the generalization itself~\cite{anonymous2025adversarial}. The main reason is that in real-world applications, reliability when faced with this type of input is crucial for decision-making tasks, preventing undesirable outcomes and potential catastrophes. In this context, previous works studied the impact of pruning on adversarial robustness and OOD generalization, supporting that pruning mechanisms enhance these metrics~\cite{Jordao:2023,Bair:2024}.

In this experiment, we aim to assess the impact of multiple structure removal promoted by our technique on adversarial attacks and OOD generalization by considering widely adopted benchmarks. To achieve this, we employ the CIFAR-C~\cite{Hendrycks:2019} and CIFAR-10.2~\cite{Lu:2020} datasets and FGSM attack to test our pruned models. Specifically, we achieve improvements in robustness even at high compression rates, for example, in CIFAR-10.2 at 81\% of FLOP reductions, there are no signs of performance degradation in ResNet110. Additionally, at a surprising compression rate of 91\%, our method enhances robustness against the FGSM attack by 2.5\%. Lastly, on CIFAR-C, all of our pruned models exhibit better robustness compared to the unpruned baseline, reaching 96\% of computational reductions with a positive delta in accuracy of almost 4\%. We observe a similar trend for other architectures such as ResNet56. These results confirm the effectiveness of our approach and highlight the potential of pruning as a mechanism capable of improving robustness. Importantly, these results confirm the applicability of our pruned models in critical scenarios.

\noindent
\textbf{GreenAI.}
Previous studies confirmed that modern models lead to high carbon emissions (CO$_{2}$) due to substantial computational power and energy consumption during both for training and deployment~\cite{Lacoste:2019,Faiz:2024,anonymous2025holistically}. Fortunately, the computational gains of our pruned models directly translate into lower CO$_{2}$ emissions. As a concrete example, on ResNet56, our pruned model with the highest FLOP reduction achieves a $76.47\%$ reduction in CO$_{2}$ emission. Another milestone is on ResNet110 by accomplishing a $83.31\%$ of CO$_{2}$ reduction. Importantly, these pruned models also improve the financial cost. In particular, we reduce $76\%$ and $83.67\%$ of these costs on ResNet56/110. We estimate these values using the Machine Learning Impact Calculator~\cite{Lacoste:2019}. In summary, we provide solid results toward GreenAI, helping to reduce CO$_{2}$ emissions and making deep models more financially accessible.

%% file: Sections/Conclusions.tex
\section{Conclusions}\label{sec:conclusions}
Modern high-performance pruning approaches exhibit promising results and distinct computational benefits in removing neurons (filters) or layers composing a model. However, few efforts exist in pruning both structures at once. In this work, we fill this gap by introducing a novel pruning approach capable of eliminating both structures at once. The central idea behind our method involves deciding which structure (neuron or layer, each one represented by a subnetwork) to eliminate during an iterative pruning process. Throughout this decision process, we confirm that choosing subnetworks that preserve the internal representation with its parent (the network that yields the subnetworks) is an effective strategy. While our modeling of this problem enables a naive solution -- random decision -- we demonstrate that employing similarity metrics, such as Centered Kernel Alignment (CKA), leads to significantly better results. In this direction, we show that our method is similarity metric-agnostic, achieving positive results with various similarity metrics.

Through experiments on standard benchmarks and architectures, we validate the effectiveness of our technique. Specifically, our method surpasses state-of-the-art pruning techniques in terms of FLOP reduction and generalization by a notable margin. In particular, at high levels of FLOP reduction, most methods face challenges in maintaining accuracy, whereas our approach either improves it or experiences only a minimal drop. Notably, we achieve milestones in FLOP reduction such as a $95.82\%$ FLOP reduction on ResNet110. Such achievements comes from the fact that our method leverages the best of both layer and neuron pruning, while existing methods are confined to a single structure.

Apart from the previous benefits, our pruned models exhibit robustness against different adversarial attacks. Finally, our models also reduce financial costs associated with training and fine-tuning and poses an important step towards GreenAI by reducing up to 83\% of carbon emissions and financial costs required for training and fine-tuning modern architectures.

In summary, given the call for efficient models, particularly in the age of foundation models, we believe our work paves the way for a new chapter in accelerating models through pruning.

\noindent
\textbf{Room for Improvement.} Despite the positive results, we believe our work offers opportunities for refinement. In this sense, one could incorporate more pruning settings as additional branches in a multi-branched decision tree. 

%% file: Sections/Acknowledgments.tex
\section*{Acknowledgments}
This study was financed, in part, by the São Paulo Research Foundation (FAPESP), Brasil. Process Number \#2023/11163-0. 
This study was financed in part by the Coordenação de Aperfeiçoamento de Pessoal de Nível Superior – Brasil (CAPES) – Finance Code 001. 
The authors would like to thank grant \#402734/2023-8, National Council for Scientific and Technological Development (CNPq). 
Artur Jordao Lima Correia would like to thank Edital Programa de Apoio a Novos Docentes 2023. Processo USP nº: 22.1.09345.01.2. 
Anna H. Reali Costa would like to thank grant \#312360/2023-1 CNPq.

%% file: Sections/Appendix.tex
\section{Appendix}\label{sec:app}
\subsection{Robustness to Pruning Criteria}\label{sec:ablation}
In this experiment, we highlight the potential of our method by illustrating its flexibility w.r.t popular pruning criteria $c$. Here, we prune ResNet32 using $\ell_1$-norm (L1) and compare it with the Kullback-Leibler (KL). While preserving generalization performance, L1 achieves a FLOP reduction of $54.60\%$, and KL achieves $61.99\%$. From these results, we confirm that our method successfully works with others criteria. 

\subsection{Results on Shallow Architectures}\label{sec:shallow}
In this experiment, we demonstrate the effectiveness of our method on shallow architectures, specifically ResNet32/44. Table~\ref{tab:resnet32:resnet44} summarizes the results. From Table~\ref{tab:resnet32:resnet44}, we highlight that the our method reduces $61.99\%$ and $71.31\%$ of FLOPs respectively from architectures ResNet32/44, without degrading the model accuracy. On ResNet32, we achieve a milestone of $96.15\%$ FLOPs reduction, surpassing any other reports from previous work~\cite{He:2023}. This behavior aligns with our analysis of deeper models and reinforces the effectiveness of our method on shallow architectures.
	
Overall, our method accomplishes notable results, outperforming state-of-the-art pruning methods. As a clarification, the table shows few methods due to the absence of pruning studies reporting results on these architectures~\cite{He:2023,Cheng:2024}. 
	
	\begin{table}[!b]
		\centering
		\footnotesize
		\addtolength{\tabcolsep}{-3pt}
		\renewcommand{\arraystretch}{1.0}
		\caption{Comparison of state-of-the-art methods on CIFAR-10.}
		\label{tab:resnet32:resnet44}
		\begin{tabular}{llrc}
			\hline
			& Method                             & $\Delta$ Acc.     & FLOPs (\%)     \\ \hline
		\multicolumn{1}{l|}{}                          & DAIS~\cite{Guan:2023} (TNNLS, 2023)       & (+) 0.57 & 53.90 \\
			\multicolumn{1}{l|}{} & SOKS~\cite{Liu:2023} (TNNLS, 2023) & (-) 0.80          & 54.58          \\
			\multicolumn{1}{l|}{} & Pons et al.~\cite{Pons:2024}                         & (+) 0.05          & 54.61 \\ \multicolumn{1}{l|}{\multirow{2}{*}{ResNet32}}		&		\textbf{Ours}           & \textbf{(+) 0.71}         & \textbf{57.40}      \\
			\cline{2-4} 
			\multicolumn{1}{l|}{} & Pons et al.~\cite{Pons:2024}                         & (-) 0.18 & 61.44 \\
			\multicolumn{1}{l|}{}		&		\textbf{Ours}           & \textbf{(+) 0.09}         & \textbf{61.99}      \\ \cline{2-4}
			\multicolumn{1}{l|}{}		&		\textbf{Ours}           & (-) 2.83         & 87.06      \\
			\multicolumn{1}{l|}{}		&		\textbf{Ours}           & (-) 7.63         & \textbf{96.15}      \\
			\hline
			\cline{2-4} 
			\multicolumn{1}{l|}{} & Pons et al.~\cite{Pons:2024}                         & (+) 0.22          & 62.95          \\
			\multicolumn{1}{l|}{\multirow{1}{*}{ResNet44}}		&		\textbf{Ours}           & \textbf{(+) 0.66}         & \textbf{71.31}      \\
			\cline{2-4} 
			\multicolumn{1}{l|}{}		&		\textbf{Ours}           & (-) 1.26         & \textbf{83.19}      \\
			\hline
		\end{tabular}
	\end{table}
	
\subsection{Results on the Transformer Architecture}\label{sec:transformer}
This experiment focuses on demonstrating the potential of our method beyond ResNet architectures. As prior studies indicate, pruning techniques extend to modern self-attention architectures, such as Transformers~\cite{Cheng:2024}. In this work, we show that our method is compatible with this architecture. To evaluate its efficiency, we alternate structural pruning between layers and heads in the multi-head attention blocks. We conduct these experiments using tabular data~\cite{Sena:2021}.
	
Our original, unpruned, Transformer architecture consists of $10$ layers, each featuring 128 heads with projection dimensions of $64$. We train this architecture for $100$ epochs on each dataset before pruning it using the proposed method.
	
\begin{figure}[h]
		\centering
		\includegraphics[width=0.23\linewidth]{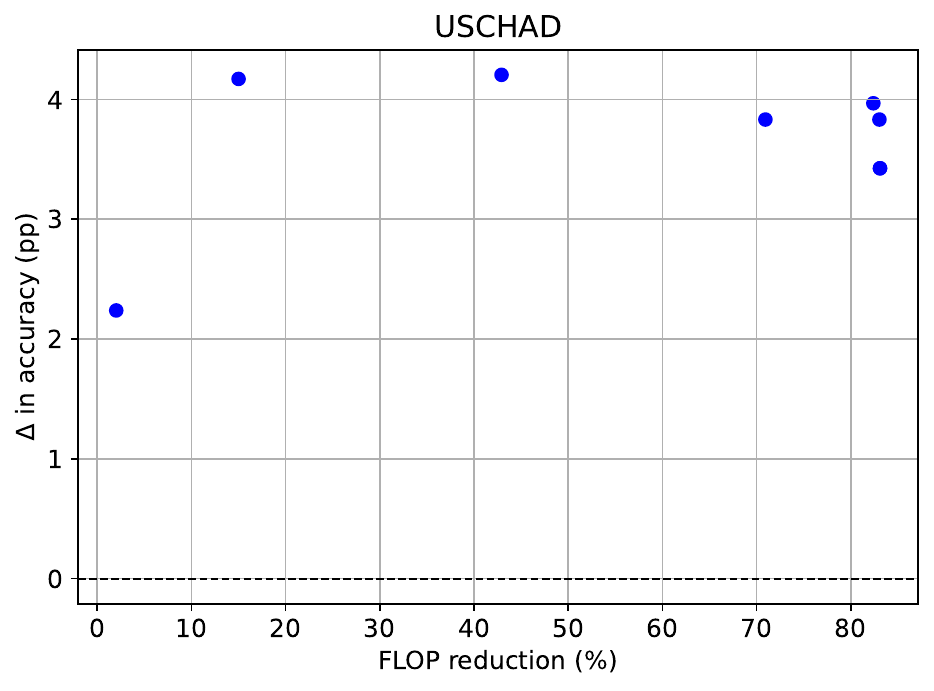}
		\includegraphics[width=0.23\linewidth]{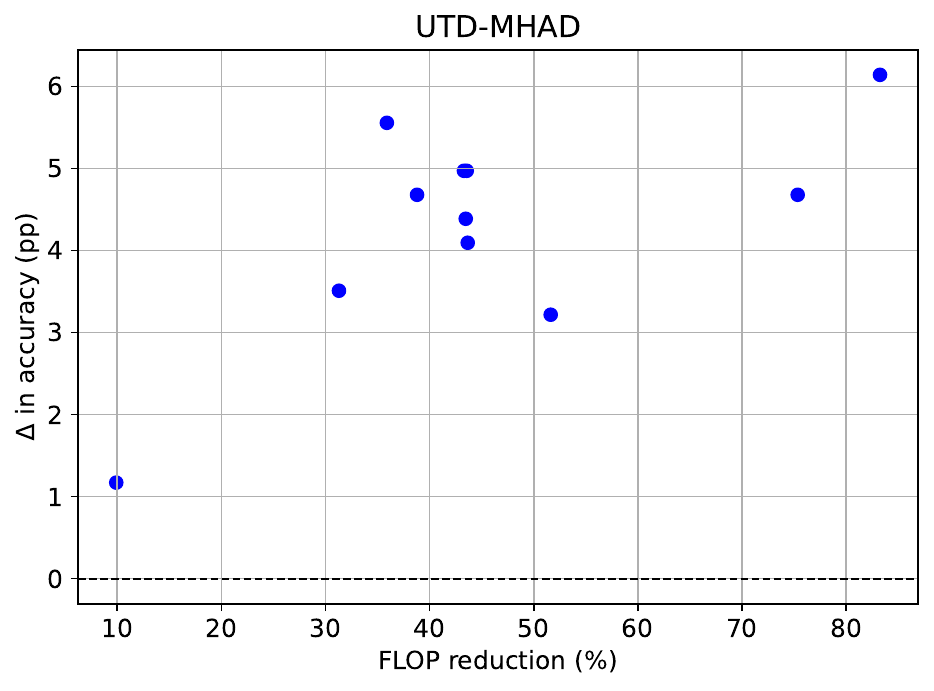}
		\includegraphics[width=0.23\linewidth]{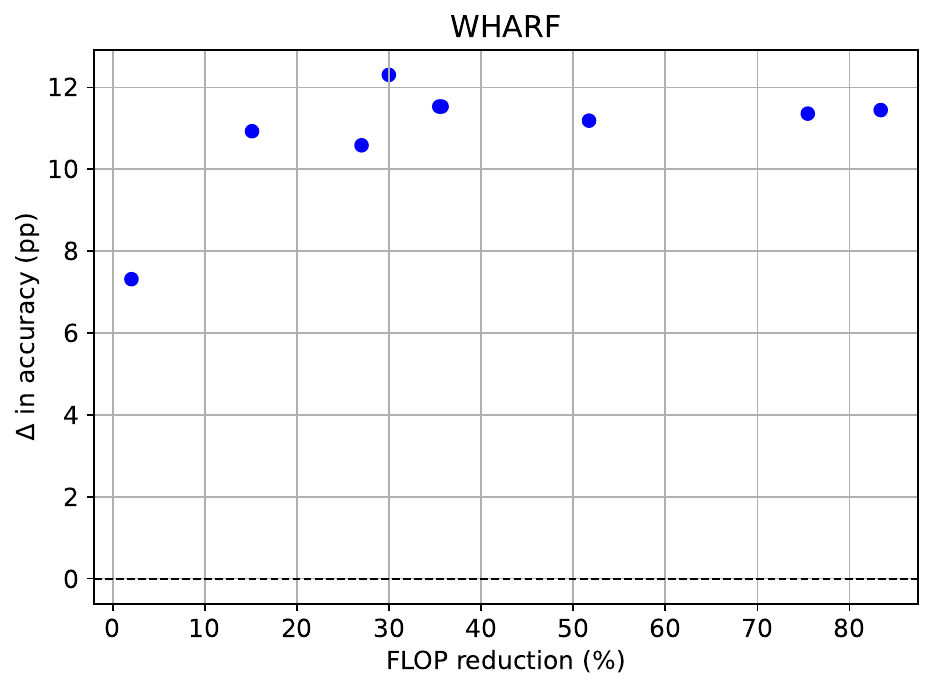}
		\includegraphics[width=0.23\linewidth]{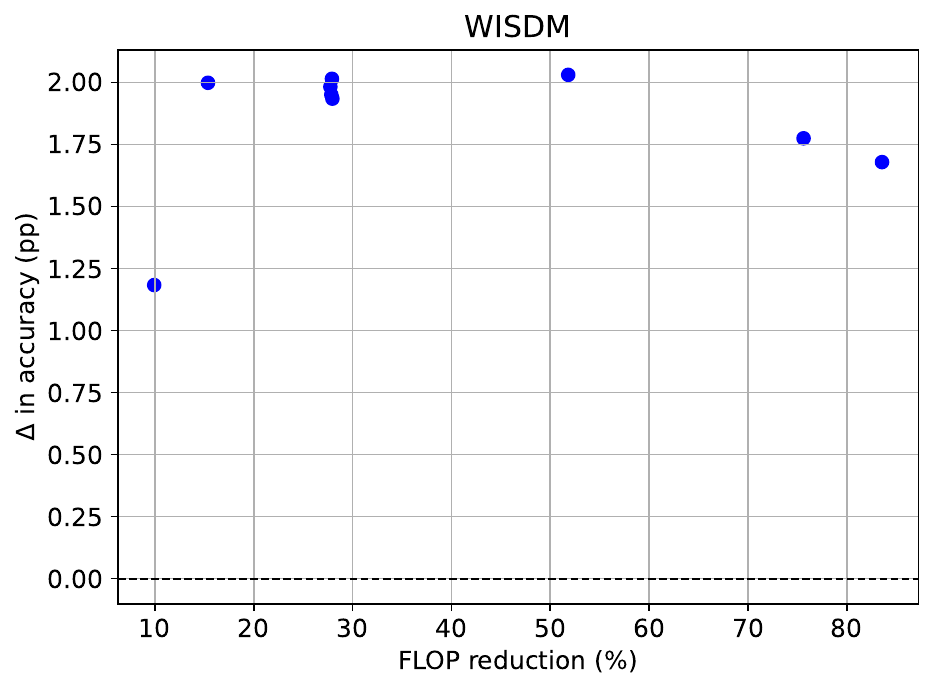}
		\caption{Performance of our method pruning Transformer architectures.} 
		\label{fig:transformer_results}
	\end{figure}

Figure~\ref{fig:transformer_results} shows the results, where each blue circle represent a pruned model and an improvement in accuracy. According to the figure, all subnetworks show gains in accuracy. We intend to explore this behavior further in future research on LLMs.
	
We also demonstrate that our method, when pruning Transformers, is effective even when changing the pruning criterion. As a concrete example, we prune the model using KL and L1, achieving respectively $83.91\%$ and $85.08\%$ of FLOP reduction, while maintaining generalization. 
	
Overall, the findings of this experiment shows a breakthrough compared to structural pruning on ResNet architectures and Figure~\ref{fig:transformer_results} confirms that our method applies effectively to Transformers for tabular data.